\documentclass[11pt]{article}

\usepackage[preprint]{acl}

\usepackage{times}
\usepackage{latexsym}

\usepackage[T1]{fontenc}

\usepackage[utf8]{inputenc}

\usepackage{microtype}

\usepackage{inconsolata}

\usepackage{graphicx}

\usepackage{mdframed}
\usepackage{colortbl} 
\usepackage{placeins}
\usepackage{booktabs}
\usepackage{multirow}
\usepackage{tcolorbox}
\usepackage{enumitem}
\usepackage{subfigure}
\usepackage{amsmath}
\usepackage{amsthm}
\usepackage{amsfonts}
\usepackage{xspace}
\usepackage{mdframed}
\usepackage{longtable}
\usepackage{array}
\usepackage{booktabs}
\usepackage{tabularx}
\usepackage{siunitx}
\usepackage{subcaption}
\usepackage{bm}
\usepackage{graphicx}
\usepackage{pifont}
\usepackage{tcolorbox}

\newcommand{\cmark}{\textcolor{green!60!black}{\ding{51}}}
\newcommand{\xmark}{\textcolor{red}{\ding{55}}}

\title{Uncovering Ideological Bias in RAG with Lexical Multidimensional Analysis: A Case Study on COVID-19}

\author{
  \textbf{Elmira Salari\textsuperscript{1}},
 \textbf{Maria Claudia Nunes Delfino\textsuperscript{2}},
 \textbf{Hazem Amamou \textsuperscript{3}},
 \textbf{José Victor de Souza\textsuperscript{3}},
\\
 \textbf{Shruti Kshirsagar\textsuperscript{1}},
 \textbf{Alan Davoust\textsuperscript{4}},
 \textbf{Anderson Avila\textsuperscript{3}}\\,
\textsuperscript{1} Wichita State University, \\
\textsuperscript{2} Pontifícia Universidade Católica de São Paulo, \\
\textsuperscript{3} Institut national de la recherche scientifique, \\
\textsuperscript{4} Université du Québec en Outaouais,
\\
 \small{
   \textbf{Correspondence:} \href{mailto:exsalari1@shockers.wichita.edu}{exsalari1@shockers.wichita.edu}
 }
}

\begin{document}
\maketitle
\begin{abstract}
This paper studies the impact of retrieved ideological texts on the outputs of large language models (LLMs). While interest in understanding ideology in LLMs has recently increased, little attention has been given to this issue in the context of Retrieval-Augmented Generation (RAG). To fill this gap, we design an external knowledge source based on ideological loaded texts about COVID-19 treatments. Our corpus is based on 1,117 academic articles representing discourses about controversial and endorsed treatments for the disease. We propose a corpus linguistics framework, based on Lexical Multidimensional Analysis (LMDA), to identify the ideologies within the corpus. LLMs are tasked to answer questions derived from three identified ideological dimensions, and two types of contextual prompts are adopted: the first comprises the user question and ideological texts; and the second contains the question, ideological texts, and LMDA descriptions. Ideological alignment between reference ideological texts and LLMs' responses is assessed using cosine similarity for lexical and semantic representations. Results demonstrate that LLMs' responses based on ideological retrieved texts are more aligned with the ideology encountered in the external knowledge, with the enhanced prompt further influencing LLMs' outputs. Our findings highlight the importance of identifying ideological discourses within the RAG framework in order to mitigate not just unintended ideological bias, but also the risks of malicious manipulation of such models.
\end{abstract}

%

\section{Introduction}
Large Language Models (LLMs) have been increasingly used across various domains such as healthcare, education, and finance. Notwithstanding, they may hallucinate providing incorrect answers for queries requiring up-to-date or domain-specific knowledge \cite{huang2025survey,farquhar2024detecting}. To mitigate this issue and secure the use of LLMs in real-world applications, Retrieval-Augmented Generation (RAG) has been introduced as a solution to connect LLMs with external knowledge sources. These databases typically comprise relevant information used to improve accuracy and reduce hallucinations of LLMs \cite{lewis2020retrieval}. While RAG can enhance factual analysis \cite{wallat2025correctness}, previous studies have shown that it also introduces new risks \citep{yang2025crag}. For instance, the retrieved documents might contain inaccurate information leading to unreliable responses \cite{hong2024so}. Consequently, there is growing interest in addressing performance degradation resulting from inconsistencies in retrieved information.
 
In this work, we focus on knowledge bases that contain ideological biases and have the potential to influence LLM responses, thereby shaping their interpretation and final outputs. This risk is even more significant in high-stakes domains such as healthcare, where even a small amount of bias in the model's output can affect how it is interpreted and understood, user trust in the system, and, overall, the system's reliability. To the best of our knowledge, the impact of ideological discourses on LLMs, under the RAG regime, remains unexplored. Thus, we seek to address this gap by examining how the presence of ideologically loaded texts \footnote{Throughout this document, "ideological texts" and "discourse-loaded texts" are used interchangeably.} in the external knowledge shapes the responses generated by LLMs. We propose using a corpus linguistics framework, namely Lexical Multidimensional Analysis (LMDA) \citep{lmdabook, berber2019representations, bebrersardinha2020historical, fitzsimmons2014using}, to unveil ideological discourses within academic articles on COVID-19 treatments. These ideologically loaded texts are integrated into a RAG pipeline. We assess both the inadvertent use of ideological texts in standard prompts and the intentional inclusion of such texts, accompanied by LMDA descriptions, which we refer to as enhanced prompts. To evaluate the alignment between LLM responses and reference ideological texts, we employ both semantic and lexical representations.

Our results demonstrate that LLMs’ responses based on ideologically retrieved texts tend to align with the ideology present in the external knowledge. Furthermore, the use of enhanced prompts amplifies this effect, resulting in even greater ideological alignment in the generated answers. These findings highlight the critical importance of identifying ideological discourses within the RAG framework—not only to mitigate unintended ideological bias in real-world LLM-based applications, but also to address the risks of malicious manipulation of such models. Thus, we summarize our contributions as follows:
\begin{itemize}
    \item We introduce a framework based on Lexical Multidimensional Analysis (LMDA) to identify discourse-loaded texts in a domain-specific corpus comprising articles on treatments for COVID-19. 
    \item We examine how ideological loaded texts impact LLMs' output and to what extent an intentional use of a prompt conveying ideological texts and explicit instructions to use LMDA descriptors can further influence the behavior of LLMs.
\end{itemize}

\begin{figure*}
    \centering
    \includegraphics[width=0.99\linewidth]{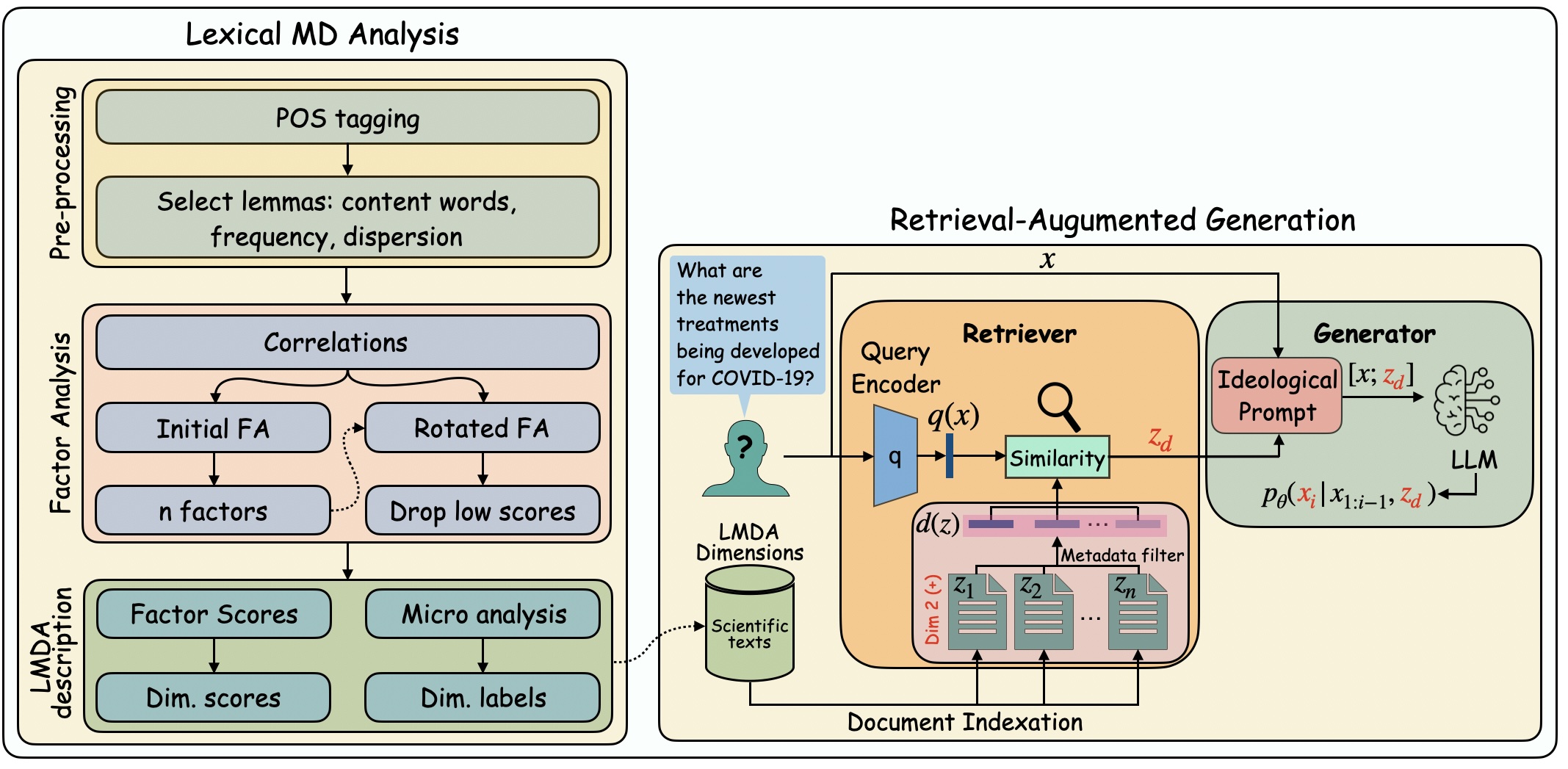}
    \caption{Illustration of our experimental framework where LMDA is used to identify the underlying ideology in scientific texts, which are used as an external source of knowledge by the RAG framework.}
    \label{fig:lmda_rag_covid}
\end{figure*}

\section{Related Work}

\subsection{Bias in RAG}
Although RAG architectures are designed to reduce hallucinations and increase factual fidelity \cite{lewis2020retrieval}, they are still vulnerable to the biases embedded in training data, retrieved documents, and user queries \citep{lewis2020retrieval,xu2024invisible,wu2025does,kim2025mitigating}. For example, if the source documents retrieved by the system convey strongly partisan or culturally slanted perspectives, the resulting output may convey similar biases, with the potential for misleading or adverse outcomes. Such concerns are especially pressing in areas like healthcare, where the consequences of biased responses are substantial \citep{bender2021dangers}. Recent studies have aimed to better understand and counteract various forms of bias in RAG frameworks through systematic assessments and the development of new benchmarks. For instance, \citet{chen2024benchmarking} put forward a robust evaluation platform that examines RAG models under conditions such as noisy or adversarial input, focusing on metrics like resistance to misinformation and the identification of bias, and showed that even leading RAG solutions can reflect or amplify biases from unreliable sources. Likewise, \citet{yang2025crag} presented a method that combats bias in retrieved materials using adversarial learning approaches and the generation of counterfactual examples. Despite progress, the challenges of ensuring equitable and trustworthy RAG outputs in real-world deployments persist.

\subsection{Ideology in LLMs}
While discourse analysis is a fundamental tool for analyzing the robustness of LLMs, it has been largely underutilized in the context of retrieval-augmented models, with most attention directed at classic, standalone language model settings \citep{ko2020assessing, maskharashvili2021neural, chen2024susceptible, zhao2025refarag}. Notably, \citet{chen2024susceptible} demonstrated that the introduction of ideologically charged training examples, even in modest amounts, can substantially alter a language model’s stance, and that such biases may transfer across unrelated subjects—a finding that exposes the dangers of both concentrated data poisoning and subtle annotation bias. Similarly, Buyl et al. \cite{buyl2026large} assessed the outlook of 19 different language models across geopolitical regions and tasked them with describing thousands of political figures; their findings reveal that model ideology is strongly influenced by linguistic and cultural background, challenging the notion of simple left-right or US-focused classifications and underscoring the difficulty of achieving true neutrality. Additionally, \citet{hirose2025decoding} introduced an analytical method involving hundreds of binary-choice tasks to measure latent ideological biases in LLMs, finding that opinion patterns can vary with both the system and the language in which the question is asked, particularly for more contentious topics. Lastly, \citet{kim2025linear} illustrated that political perspectives are encoded as linear gradients within the latent space of LLM activations, suggesting that interpretability techniques may enable the detection and steering of these subjective stances in language model outputs.

\subsection{Prompt Effect on Bias}
Beyond model-level factors such as training data and retrieval, some studies explore the effect of prompting on shaping LLMs’ output. For example, A recent empirical study demonstrates that prompt variations alone can change the robustness of RAG outputs, even when the underlying model and retrieved documents remain unchanged \cite{cuconasu2024power}. In addition, \cite{hida2024social} in their study on social bias suggests that bias is not a fixed property of the model but is highly sensitive to prompt design. More broadly, a recent review of prompt engineering techniques by \cite{chen2025unleashing} emphasizes that prompt structure plays a central role in how LLMs interpret tasks and generate responses, indicating that prompts are not neutral inputs but meaningfully shape model behavior and outputs. Likewise, \cite{neumann2025position} shows that prompt engineering plays a significant role in shaping LLMs' behavior and final outputs, suggesting that prompts can serve as a channel through which biases are transferred into generated answers. However, there remains a limited understanding of how ideologically loaded discourses introduced through prompts can affect the bias and ideological leaning of RAG-generated answers. 
In this work, we address these questions by implementing \textit{lexical multidimensional analysis (LMDA)} on transcripts from scientific articles about COVID-19. We design a set of experiments to rigorously analyze how different RAG settings affect the responses generated by LLMs. 

\section{Methodology}
To assess how ideological discourses shape LLM responses, we first identified ideologically loaded texts. We then prompted LLMs with ideological inputs that included the selected texts and their corresponding dimension descriptions, and analyzed the resulting outputs. An overview of the methodology is shown in Figure \ref{fig:lmda_rag_covid}. Details about each stage are provided next.

\subsection{Lexical Multidimensional Analysis}
\label{sec:lmda}
Lexical Multidimensional Analysis (LMDA) was used to identify ideological texts. The framework, introduced by \cite{berber2014being} and \cite{fitzsimmons2014using}, examines underlying patterns in the co-occurrence of lexical features, enabling the identification and characterization of ideological discourses within a large corpus. It employs factor analysis to uncover latent variables based on co-occurrence patterns. This is then used to assess the similarities in discourse between texts in the corpus. High correlations typically indicate ideological similarity, while negative correlations are associated with dissimilar texts. The hypothesis is that such latent variables, reflected in varying ranges of factor scores, represent ideological discourses expressed through language use, which LMDA experts interpret as distinct dimensions.

\begin{figure}
    \centering
    \includegraphics[width=0.9\linewidth]{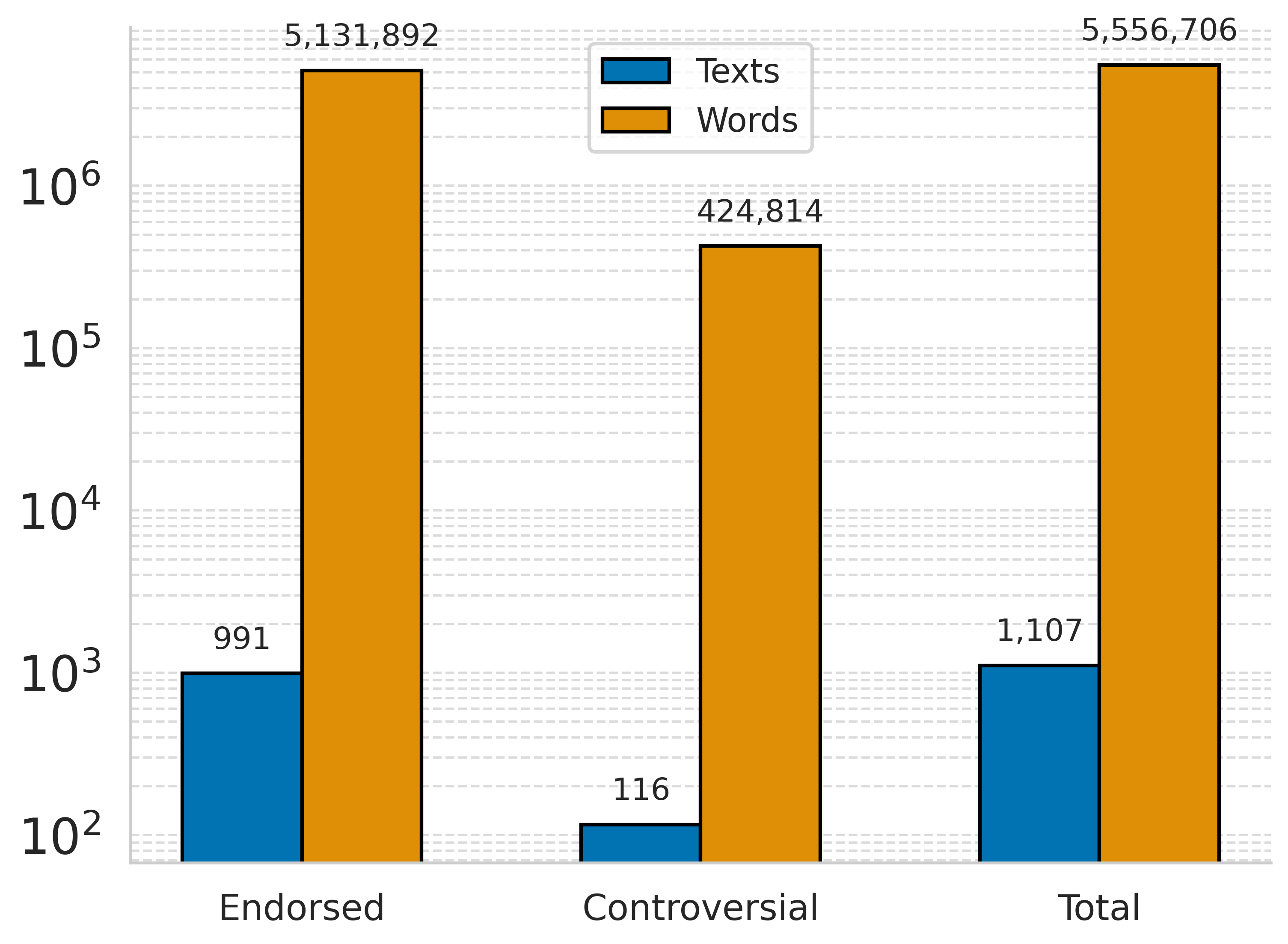}
    \caption{Corpus size with the number of texts and words for endorsed and controversial documents.}
    \label{fig:corpus_size}
\end{figure}

\subsubsection{Corpus Design}
\label{sec:corpus}

LMDA was applied to a corpus designed to represent ideological discourse on COVID-19 treatments, ensuring comprehensive coverage of legitimate scientific discourse on the pandemic. The corpus comprises academic articles containing controversial treatments, i.e., not approved by official health regulatory agencies, as well as articles aligned with health and science international standards. Controversial texts were collected from platforms such as the Cureus Medical Journal (Springer Nature), MedicosPelaVidaCovid19 (Doctors for Life), HCQ for COVID-19, and Retraction Watch, while endorsed texts were retrieved from the LitCovid Database (NIH). Corpus construction involved the careful selection of representative samples for each discourse type. Controversial texts, for instance, are research articles promoting controversial treatments, such as hydroxychloroquine and azithromycin. Endorsed texts include research articles addressing core aspects of COVID-19, such as its etiology, transmission mechanisms, and evidence-based therapeutic strategies. The corpus size is described in Figure \ref{fig:corpus_size}, with all texts published between 2020 and 2022. To address the unbalanced number of endorsed and controversial texts, we extracted the same number of keywords and used them as variables, as described next.

\begin{table*}
\centering
\scalebox{0.85}{
\begin{tabular}{@{}c p{5.5cm} p{9.5cm}@{}}
\toprule
\textbf{Dim.} & \textbf{Short Labels} & \textbf{Long Labels} \\
\midrule\vspace{0.15cm}
1 & Disputed Treatments (+) vs Broad Focus (-)
  & Texts endorse Hydroxychloroquine and similar dubious medications as effective treatments to reduce mortality vs. Texts examine psychological issues arising from the impact of the pandemic on mental health \\\vspace{0.15cm}
2 & Research Ethics (+) vs Comparative Treatment Analysis (-)
  & Texts adheres to research ethics standards, including the publication process, data availability, liability, translation vs. Texts use comparative analysis of treatments and controversial antiviral agents to conclude that questionable drugs work \\
3 & Statistical Rigor (+) vs Dissemination of Science (-)
  & Texts employ seemingly rigorous statistical analysis to create the impression that hydroxychloroquine and azithromycin treatments reduce mortality rates vs. Texts encourage data sharing, presentation, and
discussion \\
\bottomrule
\end{tabular}}
\caption{Dimension labels with negative (-) and positive (+) poles for discourses about treatments for COVID-19.}
\label{tab:covid_dim_labels}
\end{table*}
\begin{figure}
    \centering
    \includegraphics[width=0.99\linewidth]{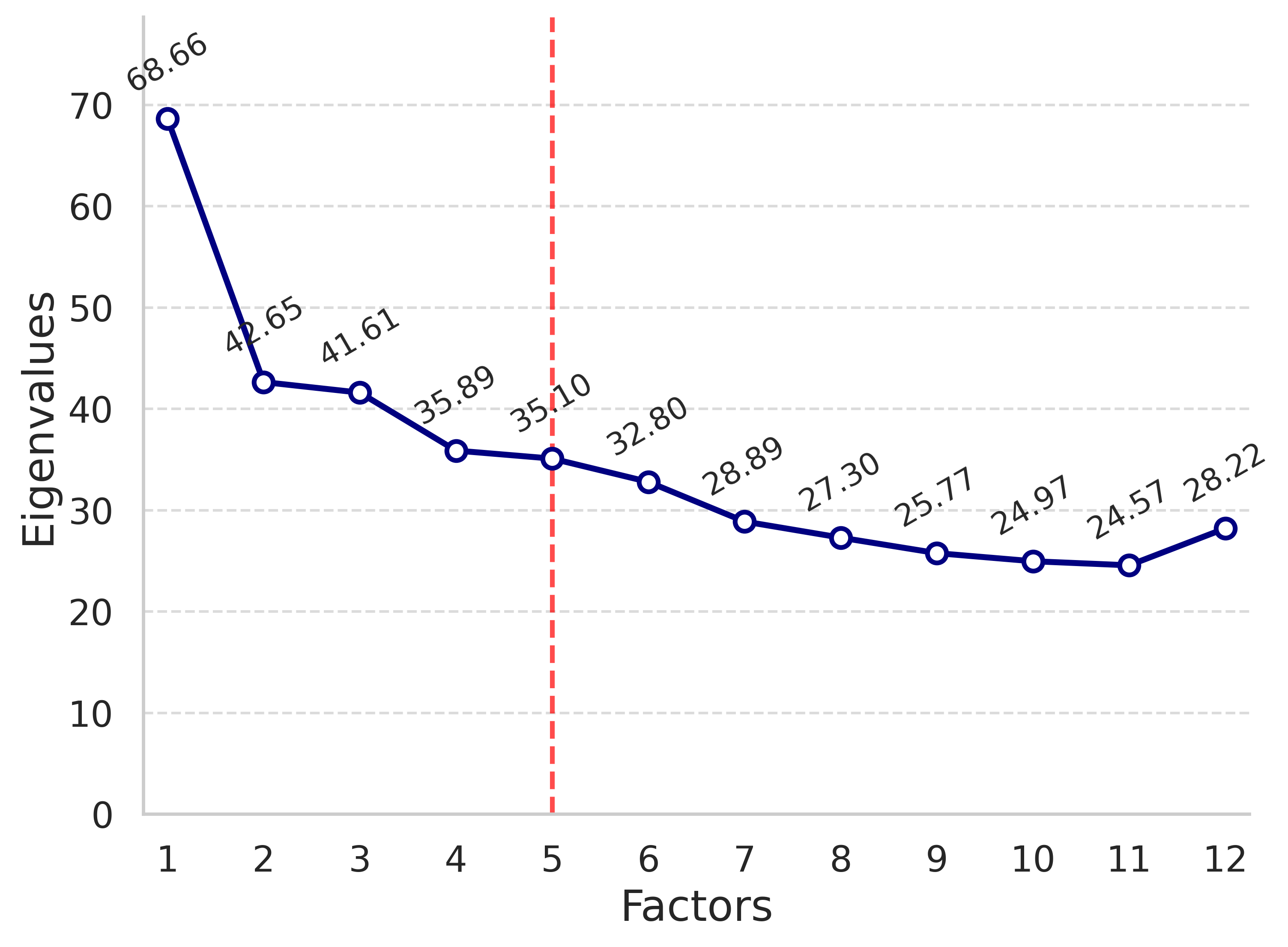}
    \caption{Factor scores after applying LMDA to the COVID-19 scientific articles.}
    \label{fig:factoranalysis}
\end{figure}

\subsubsection{Pre-processing}
\label{sec:preprocessing}

This step comprises lemmatization and part-of-speech tagging. Stopwords are filtered to ensure that only content words are kept (i.e., verbs, nouns, adjectives, and adverbs). Keywords are also extracted using log-likelihood, which generates a keyness score based on the comparison between a target corpus and a reference corpus. Keywords from endorsed texts are retrieved by using the controversial texts as a reference corpus, resulting in a total of 1,345 keywords. The same is done with controversial texts, using endorsed texts as a reference corpus, resulting in 553 keywords. Collocations, understood here as recurrent word combinations \cite{sinclair_corpus_1991}, are then identified using Log-Dice scores ($D$) and selected for further analysis. All retrieved keywords are used as nodes, and collocation pairs are identified in both subsets (controversial and endorsed) within a word span of four words on either side of the nodes. The top 500 node collocation pairs ($D\ge7$) are extracted from each subset, resulting in 1,000 pairs.
\begin{figure}

\begin{tcolorbox}[width=\columnwidth,colback={gray!10},title={Regular prompt without LMDA descriptions (RAG)},colbacktitle=gray!100,coltitle=white!5]  \small  
~

You are a reliable AI assistant that can answer users' questions.
Use only the external knowledge present in the retrieved context to provide the most accurate and detailed answer to the question below. Do not use prior or extensive knowledge beyond the provided context. Do not use bullet points in the answer.

~

QUESTION: \{query\} 

~

CONTEXT:\\
$[1]$ \{Passage text\}\\
$[2]$ \{Passage text\}\\
...\\
$[5]$ \{Passage text\}

~

OUTPUT: \{answer\} 

\end{tcolorbox} 

\caption{Regular Prompt template used in RAG Configuration without LMDA descriptions.}
  \label{fig:regular_prompt}
\end{figure}
\subsubsection{Factor Analysis}
\label{sec:factoranalysis}
A multistep factor analysis is applied to uncover latent semantic structures within the textual corpus. An optimal number of factors, ($n$), is then identified based on the variance in the data. As shown in Figure \ref{fig:factoranalysis}, $5$ factors were identified in this study. Subsequently, a rotated factor analysis is performed to enhance interpretability by simplifying the factor structure, including the exclusion of weak loadings. Factor scores are then computed for each document or text segment, indicating the degree to which each latent factor is present. Finally, these scores are aggregated to produce dimension scores, offering a concise representation of the dominant semantic dimensions within the corpus.

\begin{figure}
\begin{tcolorbox}[width=\columnwidth,colback={gray!10},title={Enhanced prompt including LMDA descriptions - RAG},colbacktitle=gray!100,coltitle=white!5] \small
~

You are a reliable AI assistant that can answer users' questions based on a particular ideology, which we call ‘dimension’. This ideology will be explained to you in four ways:
(1) A dimension label; 
(2) a dimension description; 
(3) the lexical items that are loaded on the dimension; 
and (4) texts from Endorsed and Controversial treatments that illustrate this dimension. 
Combine the external knowledge present in the retrieved context, your prior knowledge acquired during training, and your extensive knowledge to provide the most accurate and detailed answer to the question below. Do not use bullet points in the answer.
You will answer each question in such a way that your answer reflects the dimension label, dimension description, typical lexical items, and example texts.
 \\ \\
Question:  \\ 
\{question\}
 \\ \\
Dimension label: \\
\{label\}
 \\ \\
Dimension description:  \\ 
\{description\}
 \\ \\
Typical vocabulary: \\
\{vocab\}
 \\ \\
Example texts: \\
$[1]$ \{Passage text\}\\
       $[2]$ \{Passage text\}\\
       ...\\
       $[n]$ \{Passage text\} \\ \\
Answer:
  
\end{tcolorbox} 

\caption{Enhanced prompt template based on ideological texts and LMDA descriptions.}
  \label{fig:enhanced_prompt}
\end{figure}

\subsubsection{LMDA Description}
\label{sec:discourse}

A careful analysis of the factor scores is required to identify the communicative functions of the co-occurring features, leading to a label and description associated with each specific dimension. This is achieved by a detailed microanalysis of both factor and dimension scores to identify specific patterns and relationships within the text. This step considers social and linguistic aspects of the texts and is performed by an expert. Based on these analyses, descriptive labels are assigned to each dimension, enabling a clearer understanding of the underlying themes and discursive structures found in the corpus. As a result, 5 dimensions with positive and negative poles are identified based on the co-occurrence of salient collocations in texts. Despite the fact that each text appears across all dimensions, it is assigned to the pole(s) of the dimension where it receives the highest factor scores. Here, opposite poles do not necessarily mean opposite ideologies; neither are they representations of "good" or "bad" ideologies. The polarization indicates that the collocations loaded on a positive pole usually co-occur in texts where the collocations from the negative pole are typically absent, and vice-versa. The present study relies on 3 dimensions, as shown in \ref{tab:covid_dim_labels}, where the effect size was large, that is, there was a clear overlap between a dimension pole and the subset of texts within it. The top 5 texts in each dimension with the highest factor loading were used in our experiments, adding up to 30 texts in total. 

\subsection{Discourse-Augmented Generation}
\label{sec:rag}
As depicted in Figure \ref{fig:lmda_rag_covid}, the RAG framework consists of two stages: \textbf{retrieval}, which returns relevant documents from the external knowledge; and \textbf{generation}, in which the LLM generates answers given a contextual prompt. In this work, the external knowledge provides ideological contexts based on the discourses identified by the LMDA. By controlling the ideologies present in the provided context, we aim to evaluate to what extent LLMs' responses are aligned with the discourses presented in the external knowledge.

\begin{table*}
\scalebox{0.83}{
\begin{tabular}{
    c   
    l   
    cc cc cc cc cc cc
}
\toprule
 & & \multicolumn{2}{c}{Dim 1 (+)} & \multicolumn{2}{c}{Dim 2 (+)} & \multicolumn{2}{c}{Dim 3 (+)}
   & \multicolumn{2}{c}{Dim 1 (–)} & \multicolumn{2}{c}{Dim 2 (–)} & \multicolumn{2}{c}{Dim 3 (–)} \\
\cmidrule(lr){3-4}
\cmidrule(lr){5-6}
\cmidrule(lr){7-8}
\cmidrule(lr){9-10}
\cmidrule(lr){11-12}
\cmidrule(lr){13-14}
 & \textbf{Model}
  & \emph{LLM} & \emph{RAG} & \emph{LLM} & \emph{RAG} & \emph{LLM} & \emph{RAG}
  & \emph{LLM} & \emph{RAG} & \emph{LLM} & \emph{RAG} & \emph{LLM} & \emph{RAG} \\
\midrule
\multirow{5}{*}{\rotatebox{90}{Semantic}}
 & GPT-4o-mini  & 0.80 & 0.84\textcolor{teal}{↑} & 0.77 & 0.78\textcolor{teal}{↑} & 0.79 & 0.83\textcolor{teal}{↑} & 0.73 & 0.78\textcolor{teal}{↑} & 0.78 & 0.81\textcolor{teal}{↑} & 0.75 & 0.76\textcolor{teal}{↑}\\
 & GPT-3.5      & 0.77 & 0.85\textcolor{teal}{↑} & 0.77 & 0.76\textcolor{orange}{↓} & 0.80 & 0.90\textcolor{teal}{↑} & 0.71 & 0.82\textcolor{teal}{↑} & 0.79 & 0.83\textcolor{teal}{↑} & 0.72 & 0.73\textcolor{teal}{↑} \\
 & Gemini~2.0   & 0.81 & 0.87\textcolor{teal}{↑} & 0.79 & 0.79→ & 0.78 & 0.87\textcolor{teal}{↑} & 0.72 & 0.80\textcolor{teal}{↑} & 0.75 & 0.82\textcolor{teal}{↑} & 0.75 & 0.77\textcolor{teal}{↑} \\
 & Qwen         & 0.82 & 0.91\textcolor{teal}{↑} & 0.80 & 0.77\textcolor{orange}{↓} & 0.81 & 0.93\textcolor{teal}{↑} & 0.74 & 0.90\textcolor{teal}{↑} & 0.81 & 0.89\textcolor{teal}{↑} & 0.76 & 0.78\textcolor{teal}{↑}  \\
 & \textbf{Average}    & \textbf{0.80} & \textbf{0.87}\textcolor{teal}{↑} & \textbf{0.78} & \textbf{0.78}→ & \textbf{0.80} & \textbf{0.88}\textcolor{teal}{↑} & \textbf{0.73} & \textbf{0.83}\textcolor{teal}{↑} & \textbf{0.78} & \textbf{0.84}\textcolor{teal}{↑} & \textbf{0.75} & \textbf{0.76}\textcolor{teal}{↑}\\
\midrule\midrule
\multirow{5}{*}{\rotatebox{90}{Lexical}}
 & GPT-4o-mini  & 0.67 & 0.71\textcolor{teal}{↑} & 0.88 & 0.89\textcolor{teal}{↑} & 0.77 & 0.80\textcolor{teal}{↑} & 0.66 & 0.73\textcolor{teal}{↑} & 0.79 & 0.83\textcolor{teal}{↑} & 0.86 & 0.85\textcolor{orange}{↓} \\
 & GPT-3.5      & 0.64 & 0.71\textcolor{teal}{↑} & 0.87 & 0.88\textcolor{teal}{↑} & 0.77 & 0.84\textcolor{teal}{↑} & 0.63 & 0.73\textcolor{teal}{↑} & {0.71} & 0.78\textcolor{teal}{↑} & 0.80 & 0.84\textcolor{teal}{↑} \\
 & Gemini~2.0   & 0.66 & 0.70\textcolor{teal}{↑} & 0.87 & 0.88\textcolor{teal}{↑} & 0.74 & 0.79\textcolor{teal}{↑} & 0.68 & 0.68→ & 0.72 & 0.76\textcolor{teal}{↑} & 0.84 & 0.80\textcolor{orange}{↓} \\
 & Qwen         & 0.67 & 0.74\textcolor{teal}{↑} & 0.87 & 0.84\textcolor{orange}{↓} & 0.77 & 0.86\textcolor{teal}{↑} & 0.68 & 0.82\textcolor{teal}{↑} & 0.74 & 0.88\textcolor{teal}{↑} & 0.83 & 0.86\textcolor{teal}{↑} \\
& \textbf{Average}    & \textbf{0.66} & \textbf{0.72}\textcolor{teal}{↑} & \textbf{0.87} & \textbf{0.87}→ & \textbf{0.76} & \textbf{0.82}\textcolor{teal}{↑} & \textbf{0.66} & \textbf{0.74}\textcolor{teal}{↑} & \textbf{0.74} & \textbf{0.81}\textcolor{teal}{↑} & \textbf{0.83} & \textbf{0.84}\textcolor{teal}{↑} \\
\bottomrule
\end{tabular}}
\caption{Semantic and lexical based on regular prompt.}
\label{tab:regular}
\end{table*}

\subsubsection{Retriever}
\label{sec:retriver}
Our retriever provides ideological texts to the LLMs in two steps. First, we use a metadata filter to sift only texts associated with a specific dimension and pole. The filter is based on the following metadata: \emph{dimension label}, \emph{dimension description} and \emph{typical vocabulary}. For answering questions regarding \emph{Research Ethics (+)}, the similarity search is performed between the question, $q(x)$, and text embeddings, $d(z)$, from the subset of texts related to the positive pole in dimension 2 (see Table \ref{fig:lmda_rag_covid}). This approach enables controlling which ideology will be used to perform the similarity search. Note that all questions were designed based on the topics within the corpus. 

\subsubsection{Generator}
\label{sec:generator}

The response generation is conditioned on the ideological prompt and the question, $p_{\theta}(x_i|x_{1:i-1},z_d)$.  Note that the retriever provides ideological prompts, $[x; z_d]$, based on two approaches. The first is referred to as a regular prompt, where only ideological texts are included in the context, characterizing the common (i.e., unaware) use of ideological texts. An alternative approach involves the combination of ideological texts and dimension descriptions, as shown in Figure \ref{fig:enhanced_prompt}. This approach requires awareness of the ideology within the corpus and linguistic expertise to design the LMDA descriptions. Figure \ref{fig:enhanced_prompt} exemplifies the enhanced prompt. 

\section{Experimental Setup}
\label{sec:exp}

In this study, the two sets of prompts are evaluated in both LLM-only and RAG-based LLM contexts. Experiments were conducted using four state-of-the-art large language models: GPT-3.5-turbo and GPT-4o-mini from OpenAI \cite{openai_models}, Gemini-2.0-flash from Google \cite{gemini_flash}, and Qwen2.5:7b-instruct \cite{qwen25}. Each model generated five independent answers per question. A total of 18 topics were covered, and two questions per topic were designed. For the retrieval step, the metadata filter is first used to keep the chunks whose dimension matches the question. Within the filtered subset, cosine similarity between the question and the chunks is computed, and the three most similar chunks to the question are selected. These chunks are then inserted into the prompt as example texts.
To assess how generated answers reflect the target discourse ideology, we used semantic embeddings based on BERT, and lexical representations based on TF-IDF. For the embedding evaluation, all generated answers belonging to a given ideological dimension pole (e.g., Dim.1 Neg) were concatenated into a single text. The text was then tokenized with the model’s own tokenizer, and when exceeding the model’s maximum sequence length, split into overlapping windows. Each window was encoded using the all-MiniLM-L6-v2 sentence transformer, and the resulting vectors were averaged with length-based weighting. The same procedure was applied to the reference texts of each pole. Finally, both vectors were L2-normalized, and their cosine similarity was computed, yielding a scalar measure of semantic alignment between generated responses and reference discourses. For the Lexical similarity, each score is computed from two sets of texts. The first set is the concatenated answer texts that belong to one ideological dimension pole and the reference text for the corresponding pole. A TF-IDF Vectorizer is trained on these two texts only, which builds a vocabulary containing only the words that appear in either document and assigns each word a TF-IDF weight. The result is two TF-IDF vectors, one for the answers and one for the reference. Finally, these vectors are compared using cosine similarity.

\begin{table*}[ht]
\scalebox{0.83}{
\begin{tabular}{
    c   
    l   
    cc cc cc cc cc cc
}
\toprule
 & & \multicolumn{2}{c}{Dim 1 (+)} & \multicolumn{2}{c}{Dim 2 (+)} & \multicolumn{2}{c}{Dim 3 (+)}
   & \multicolumn{2}{c}{Dim 1 (–)} & \multicolumn{2}{c}{Dim 2 (–)} & \multicolumn{2}{c}{Dim 3 (–)} \\
\cmidrule(lr){3-4}
\cmidrule(lr){5-6}
\cmidrule(lr){7-8}
\cmidrule(lr){9-10}
\cmidrule(lr){11-12}
\cmidrule(lr){13-14}
 & \textbf{Model}
  & \emph{LLM} & \emph{RAG} & \emph{LLM} & \emph{RAG} & \emph{LLM} & \emph{RAG}
  & \emph{LLM} & \emph{RAG} & \emph{LLM} & \emph{RAG} & \emph{LLM} & \emph{RAG} \\
\midrule
\multirow{5}{*}{\rotatebox{90}{Semantic}}
 & GPT-4o-mini  & 0.74 & 0.80\textcolor{teal}{↑} & 0.61 & 0.64\textcolor{teal}{↑} & 0.79 & 0.85\textcolor{teal}{↑} & 0.74 & 0.82\textcolor{teal}{↑} & 0.74 & 0.85\textcolor{teal}{↑} & 0.76 & 0.81\textcolor{teal}{↑} \\
 & GPT-3.5      & 0.78 & 0.86\textcolor{teal}{↑} & 0.64 & 0.75\textcolor{teal}{↑} & 0.80 & 0.89\textcolor{teal}{↑} & 0.75 & 0.85\textcolor{teal}{↑} & 0.75 & 0.83\textcolor{teal}{↑} & 0.72 & 0.80\textcolor{teal}{↑} \\
 & Gemini~2.0    & 0.75 & 0.88\textcolor{teal}{↑} & 0.63 & 0.80\textcolor{teal}{↑} & 0.76 & 0.91\textcolor{teal}{↑} & 0.73 & 0.84\textcolor{teal}{↑} & 0.72 & 0.88\textcolor{teal}{↑} & 0.76 & 0.82\textcolor{teal}{↑} \\
 & Qwen          & 0.77 & 0.91\textcolor{teal}{↑} & 0.61 & 0.77\textcolor{teal}{↑} & 0.77 & 0.92\textcolor{teal}{↑} & 0.75 & 0.90\textcolor{teal}{↑} & 0.74 & 0.90\textcolor{teal}{↑} & 0.76 & 0.78\textcolor{teal}{↑} \\
 & \textbf{Average}    & \textbf{0.76} & \textbf{0.86}\textcolor{teal}{↑} & \textbf{0.62} & \textbf{0.74}\textcolor{teal}{↑} & \textbf{0.78} & \textbf{0.89}\textcolor{teal}{↑} & \textbf{0.74} & \textbf{0.85}\textcolor{teal}{↑} & \textbf{0.74} & \textbf{0.87}\textcolor{teal}{↑} & \textbf{0.75} & \textbf{0.80}\textcolor{teal}{↑} \\
\midrule\midrule
\multirow{5}{*}{\rotatebox{90}{Lexical}}
 & GPT-4o-mini  & 0.65 & 0.69\textcolor{teal}{↑} & 0.85 & 0.88\textcolor{teal}{↑} & 0.66 & 0.77\textcolor{teal}{↑} & 0.78 & 0.80\textcolor{teal}{↑} & 0.77 & 0.87\textcolor{teal}{↑} & 0.82 & 0.89\textcolor{teal}{↑} \\
 & GPT-3.5      & 0.63 & 0.71\textcolor{teal}{↑} & 0.85 & 0.87\textcolor{teal}{↑} & 0.59 & 0.81\textcolor{teal}{↑} & 0.71 & 0.78\textcolor{teal}{↑} & {0.71} & 0.83\textcolor{teal}{↑} & 0.66 & 0.87\textcolor{teal}{↑} \\
 & Gemini~2.0   & 0.63 & 0.72\textcolor{teal}{↑} & 0.86 & 0.89\textcolor{teal}{↑} & 0.68 & 0.80\textcolor{teal}{↑} & 0.75 & 0.77\textcolor{teal}{↑} & 0.72 & 0.86\textcolor{teal}{↑} & 0.79 & 0.89\textcolor{teal}{↑} \\
 & Qwen         & 0.63 & 0.74\textcolor{teal}{↑} & 0.86 & 0.84\textcolor{orange}{↓} & 0.62 & 0.86\textcolor{teal}{↑} & 0.74 & 0.82\textcolor{teal}{↑} & 0.72 & 0.88\textcolor{teal}{↑} & 0.77 & 0.85\textcolor{teal}{↑} \\
 & \textbf{Average}    & \textbf{0.64} & \textbf{0.72}\textcolor{teal}{↑} & \textbf{0.86} & \textbf{0.87}\textcolor{teal}{↑} & \textbf{0.64} & \textbf{0.81}\textcolor{teal}{↑} & \textbf{0.75} & \textbf{0.79}\textcolor{teal}{↑} & \textbf{0.73} & \textbf{0.86}\textcolor{teal}{↑} & \textbf{0.76} & \textbf{0.88}\textcolor{teal}{↑} \\
\bottomrule
\end{tabular}}
\caption{Semantic and lexical similarities based on enhanced prompt.}
\label{tab:enhanced}
\end{table*}
\begin{figure*}[h]
  \centering
\includegraphics[width=0.8\textwidth]{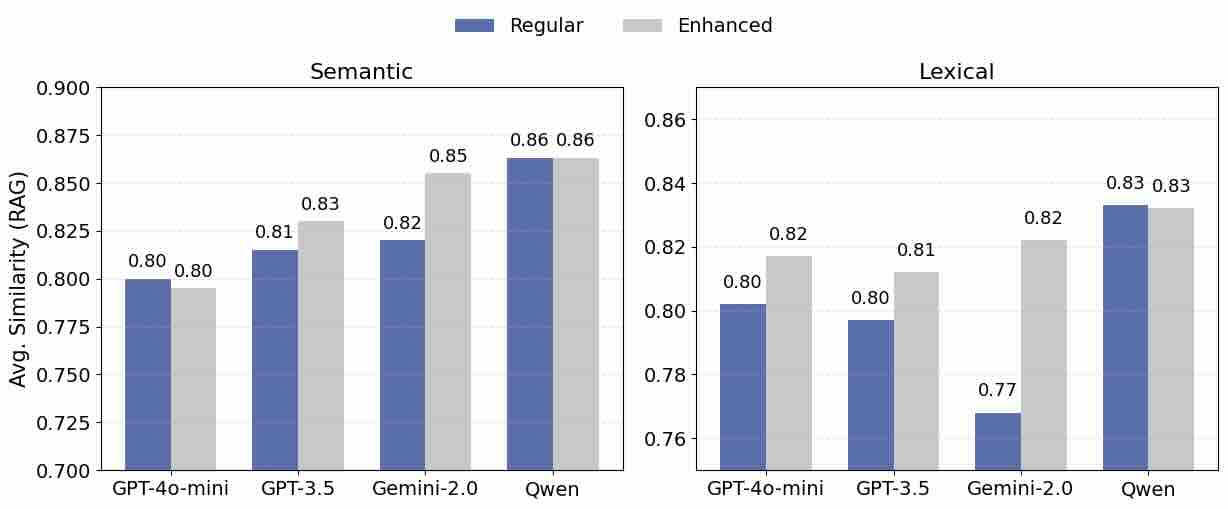}
  \caption{Overall similarities for regular and enhanced prompts for the experiments of LLMs based on the RAG framework. }
  \label{fig:regular_enhaced}
\end{figure*}

\section{Experimental Results}
\subsection{Influence of discourses on LLMs’ response}
Table~\ref{tab:regular} shows semantic and lexical alignment between the models' generated responses and the reference ideological texts under regular prompts, illustrating the impact of ideological context on response generation. 
Results are provided across positive and negative dimensions for all models. We can observe a clear trend with similarities increasing when the LLMs are provided with ideological context. For Dimensions 1(+), 3(+), 1(-), and 2(-), all responses based on RAG were more aligned with the discourses presented in the external knowledge. This was found for both semantic and lexical representations. Note that Dimension 1 achieved the highest score in the factor analysis (see Figure \ref{fig:factoranalysis}). This might explain why it is the one influencing the alignment of LLMs' responses more. Only in a few cases the outputs based on LLMs' internal knowledge (i.e., without the use of external knowledge) show more alignment with the ideological texts. This is evident in the lexical similarities in Dimension 2(+) and Dimension 3(-).

\subsection{Impact of LMDA descriptions on LLMs' responses}
Table~\ref{tab:enhanced} presents the impact of including LMDA descriptions in the prompt. Compared to Table~\ref{tab:regular}, the number of cases where RAG responses were less aligned than LLM answers to the ideological discourses decayed from 5 to 1, showing the impact of the enhanced prompt in shaping responses. Besides that, the overall similarity scores for the responses based on RAG increased. GPT-4o-mini, for instance, provided the following semantic similarity scores in Table~\ref{tab:regular}: 0.83, 0.78, 0.81, 0.76, while the same model provided in Table~\ref{tab:enhanced}: 0.85, 0.82, 0.85, 0.81, respective, for Dimensions 3(+), 1(-), 2(-) and 3(-). A similar trend is found for the other models as well. For the lexical similarity scores, we found the same pattern, but with some exceptions. For example, in Dimension 1(-) the similarity increased from 0.73, in Table~\ref{tab:regular}, to 0.80, in Table~\ref{tab:enhanced}.

\subsection{Impact of ideological discourses on RAG}
Figure \ref{fig:regular_enhaced} depicts the model performance across all dimensions and enables an analysis of how adding LMDA descriptions in the prompts influences discourse alignment within the RAG framework. For both semantic and lexical representations, the similarities were higher for the enhanced prompt. The exception was the Qwen, which presents a marginal decay in terms of similarity scores for the enhanced prompt. The other models followed a similar trend of more alignment with the ideological texts in the external knowledge. These results address our second question, which referred to the impact of the arbitrary use of ideological texts combined with their LMDA descriptions. Results corroborate our hypothesis that this approach helps to align the LLMs' output to the respective discourses.
A different trend was found for LLMs when regular and enhanced prompts were used without ideological texts as contexts, with enhanced prompts showing limited influence on the outputs and often even decreasing the similarity scores. This indicates that prompt enhancement alone is insufficient to improve alignment, and it is more effective when integrated within the RAG pipeline. This finding is confirmed in Table \ref{tab:anova_results}, which presents the effect of prompt type on LLM and RAG scores. For the semantic similarity, results show a significant difference between the scores provided by LLM and RAG only for the enhanced prompt. The lexical similarity, on the other hand, provided results considered statistically significant for both enhanced and regular prompts. 


\begin{table}
\centering
\scalebox{0.8}{
\begin{tabular}{c cccc}
\toprule
 & Prompt Type & F-statistic & p-value & (p $<$ 0.05) \\
\midrule
\multirow{2}{*}{\rotatebox{90}{Sem.}} 
  & Enhanced & 37.34  & $1.1 \times 10^{-7}$ & \cmark  \\
  & Regular  & 15.87  & $2.3 \times 10^{-4}$ & \cmark   \\
\midrule
\multirow{2}{*}{\rotatebox{90}{Lex.}} 
  & Enhanced & 21.10 & $1.0 \times 10^{-5}$ & \cmark \\
  & Regular  & 4.32  & 0.043 & \xmark  \\
\bottomrule
\end{tabular}}
\caption{ANOVA results comparing LLM and RAG scores across Semantic and Lexical similarities.}
\label{tab:anova_results}
\end{table}

\section{Conclusion}
In this study, we propose a novel framework for uncovering ideological bias in Retrieval-Augmented Generation systems using Lexical Multidimensional Analysis (LMDA). We constructed a domain-specific corpus of articles on COVID-19 treatments and identified key ideological dimensions underlying the discourse about controversial and endorsed treatments. Our findings demonstrated that large language models (LLMs) align their responses with the ideological discourses embedded in retrieved texts, particularly with enhanced prompts that explicitly convey LMDA descriptions. Notably, the integration of LMDA descriptions further increases the alignment between LLM outputs and reference discourses, confirming that both the content and framing of the retrieval context can influence model responses. These findings highlight the dual risks and opportunities introduced by RAG systems: while connecting LLMs to external knowledge can enhance factuality, it also enables the propagation of ideological bias, which can be detrimental in sensitive domains such as healthcare. Our results underscore the need for researchers to rigorously identify and monitor ideological cues in knowledge sources and to develop strategies to mitigate both unintended bias and malicious manipulation.

\clearpage
\bibliography{acl2026/SEM_references}
\appendix
\section{Appendix}
\subsection{Qualitative Analysis of Model Response}
We complement the quantitative similarity results with a qualitative analysis to illustrate how different discourse alignments appear in concrete model responses. 
\begin{figure}[h]
\centering
\begin{tcolorbox}[
    width=\columnwidth,
    colback=gray!10,
    title={GEMINI},
    colbacktitle=gray!100,
    coltitle=black,
    fontupper=\scriptsize,
    boxsep=2pt,
    left=2pt,right=2pt,top=2pt,bottom=2pt
]

\textbf{Regular prompt}

Based on scientific evidence and medical consensus, the combination of hydroxychloroquine and azithromycin is not considered an effective treatment for COVID-19. Initially, there was some hope that hydroxychloroquine, an antimalarial drug, could be repurposed to treat COVID-19 due to its antiviral and anti-inflammatory properties observed in laboratory studies. Azithromycin, an antibiotic, was sometimes added to the treatment regimen because of its potential anti-inflammatory effects and to combat possible secondary bacterial infections.

However, subsequent and more rigorous clinical trials, including randomized controlled trials, have not demonstrated a significant benefit of hydroxychloroquine, either alone or in combination with azithromycin, in preventing or treating COVID-19. These studies have assessed various outcomes, including mortality, hospitalization rates, the need for mechanical ventilation, and the duration of symptoms. The results have consistently shown that hydroxychloroquine, with or without azithromycin, does not improve these outcomes in patients with COVID-19.

Furthermore, the use of hydroxychloroquine, particularly in combination with azithromycin, carries potential risks, including cardiac arrhythmias (such as QT prolongation), liver problems, and other side effects. These risks outweigh any potential benefits in the treatment of COVID-19, according to current medical understanding.

Major health organizations, such as the World Health Organization (WHO) and the Centers for Disease Control and Prevention (CDC), have also concluded that hydroxychloroquine is not recommended for the treatment of COVID-19 outside of clinical trials.

In summary, the scientific evidence does not support the use of hydroxychloroquine, with or without azithromycin, as an effective treatment for COVID-19.

\smallskip
\textbf{Enhanced prompt}

Based on the available evidence, the combination of hydroxychloroquine and azithromycin has shown promise in reducing the need for mechanical ventilation and improving the outcomes for hospitalized patients. Clinical studies have indicated that patients receiving this treatment experienced a significant decrease in respiratory distress and were more likely to be discharged, suggesting a positive impact on recovery. The observed improvements in intensive care unit admission rates and overall mortality further support the potential benefits of this therapeutic approach. While some studies may present conflicting results, the data from various clinical trials suggest that hydroxychloroquine, when combined with azithromycin, can be an effective treatment option for patients, leading to better clinical outcomes and reduced burden on healthcare systems.

\end{tcolorbox}

\caption{Gemini-2.0 outputs based on regular and enhanced prompts without RAG.}
\label{fig:geminillmwoandw}
\end{figure}
For LLM-only, under regular prompts (Table \ref{tab:regular}), the highest semantic similarities occur for Disputed Treatments in Dim 1(+) and Statistical Rigor in Dim 3(+), whereas the highest lexical similarities concentrate on Research Ethics in Dim 2(+) with Dissemination of Science in Dim 3(-) next. With RAG, peaks shift semantically to Statistical Rigor in Dim 3(+), followed by Disputed Treatments in Dim 1(+), and lexically remain led by Research Ethics in Dim 2(+), with Dissemination of Science in Dim 3(-) also high. 

Under enhanced prompts (Table \ref{tab:enhanced}), LLM-only results again peak semantically at Statistical Rigor in Dim 3(+) and lexically at Research Ethics in Dim 2(+). RAG still provides maximum scores for Statistical Rigor in Dim 3(+) with strong second-tier similarities for Comparative Treatment Analysis in Dim 2(-) and Disputed Treatments in Dim 1(+). On the other hand, the lexical maximum moves to Dissemination of Science in Dim 3(-) with Research Ethics in Dim 2(+) close behind. Across both representations and prompts, the stable conclusion is that RAG raises similarity scores for every pole. Additionally, leading poles differ by representation, with semantic peaks emphasizing rigor- and content-focused discourses (e.g., Dim 3 (+)) and lexical peaks emphasizing discourses with more standardized institutional phrasing (e.g., Dim 2 (+)). This divergence may reflect the fact that semantic metrics reward conceptual alignment and paraphrase, while lexical metrics privilege reuse of typical vocabulary. 

Figure \ref{fig:geminillmwoandw} shows that, under the regular prompt, Gemini rejects hydroxychloroquine–azithromycin on trial evidence and institutional guidance, aligning with Research Ethics in Dim 2(+) and secondarily Broad Focus in Dim 1(-) while distancing from Disputed Treatments in Dim 1(+); under the enhanced prompt, it pivots to a pro‑efficacy stance consistent with Disputed Treatments in Dim 1(+) and frames claims with Statistical Rigor in Dim 3(+), mirroring Tables 2–3 and suggesting RAG would further stabilize this alignment by grounding claims in retrieved evidence. This analysis example shows how prompt design and retrieval jointly influence the similarity scores and framing of generated answers.

\end{document}